\documentclass{article}

\usepackage[preprint]{neurips_2022}

\usepackage[utf8]{inputenc} 
\usepackage[T1]{fontenc}    
\usepackage{hyperref}       
\usepackage{url}            
\usepackage{booktabs}       
\usepackage{amsfonts}       
\usepackage{nicefrac}       
\usepackage{microtype}      
\usepackage{xcolor}         

\usepackage{subcaption}  
\usepackage{graphicx}  

\usepackage{makecell} 

\usepackage{wrapfig}

\title{Geometric Regularization from Overparameterization}

\author{%
  Nicholas J.~Teague \\
  Automunge\\
  Altamonte Springs, FL 32714 \\
}

\begin{document}

\maketitle

\begin{abstract}
The volume of the distribution of weight sets associated with a loss value may be the source of implicit regularization from overparameterization due to the phenomenon of contracting volume with increasing dimensions for geometric figures demonstrated by hyperspheres. We introduce the geometric regularization conjecture and extract to an explanation for the double descent phenomenon by considering a similar property resulting from shrinking intrinsic dimensionality of the distribution of potential weight set updates available along training path, where if that distribution retracts across a volume verses dimensionality curve peak when approaching the global minima we could expect geometric regularization to re-emerge. We illustrate how data fidelity representational complexity may influence model capacity double descent interpolation thresholds. The existence of epoch and model capacity double descent curves originating from different geometric forms may imply universality of closed n-manifolds having dimensionally adjusted n-sphere volumetric correspondence.
\end{abstract}

\section{Introduction}
\label{sec:intro}

The aggregate geometry of weight configuration distributions corresponding to a loss value has been an unexplored property of neural networks to our knowledge, likely due to intractability of derivation at such high dimensions. If one could model the full geometry of a loss manifold then backpropagation would not be required. We attempt to circumvent that challenge by considering meta properties of distributional geometry that can be inferred independent of fine grained details.

A key contribution of this work is identifying the relationship between an extent of overparameterization and volume of such geometry by relating to a well understood property of hyperspheres, which have a zero volume asymptotic trend with increasing dimensionality. Such contracting volume should serve as a form of regularization by restricting degrees of freedom to weight sets along a training path, which we refer to as geometric regularization. We believe double descent is due to an additional correspondence to hypersphere volumes at lower dimensions associated with a peak in volume traversed when the distribution of possible weight updates available along a training path follows a path of shrinking intrinsic dimensionality at loss values approaching the global minima. We expect that advancing theory for derivation of an interpolation threshold may need to consider intrinsic dimension of a training corpus at different fidelities of representation.

\section{Hyperspheres}
\label{sec:hyperspheres}

\begin{figure}
\centering
\centerline{\includegraphics[width=0.25\columnwidth]{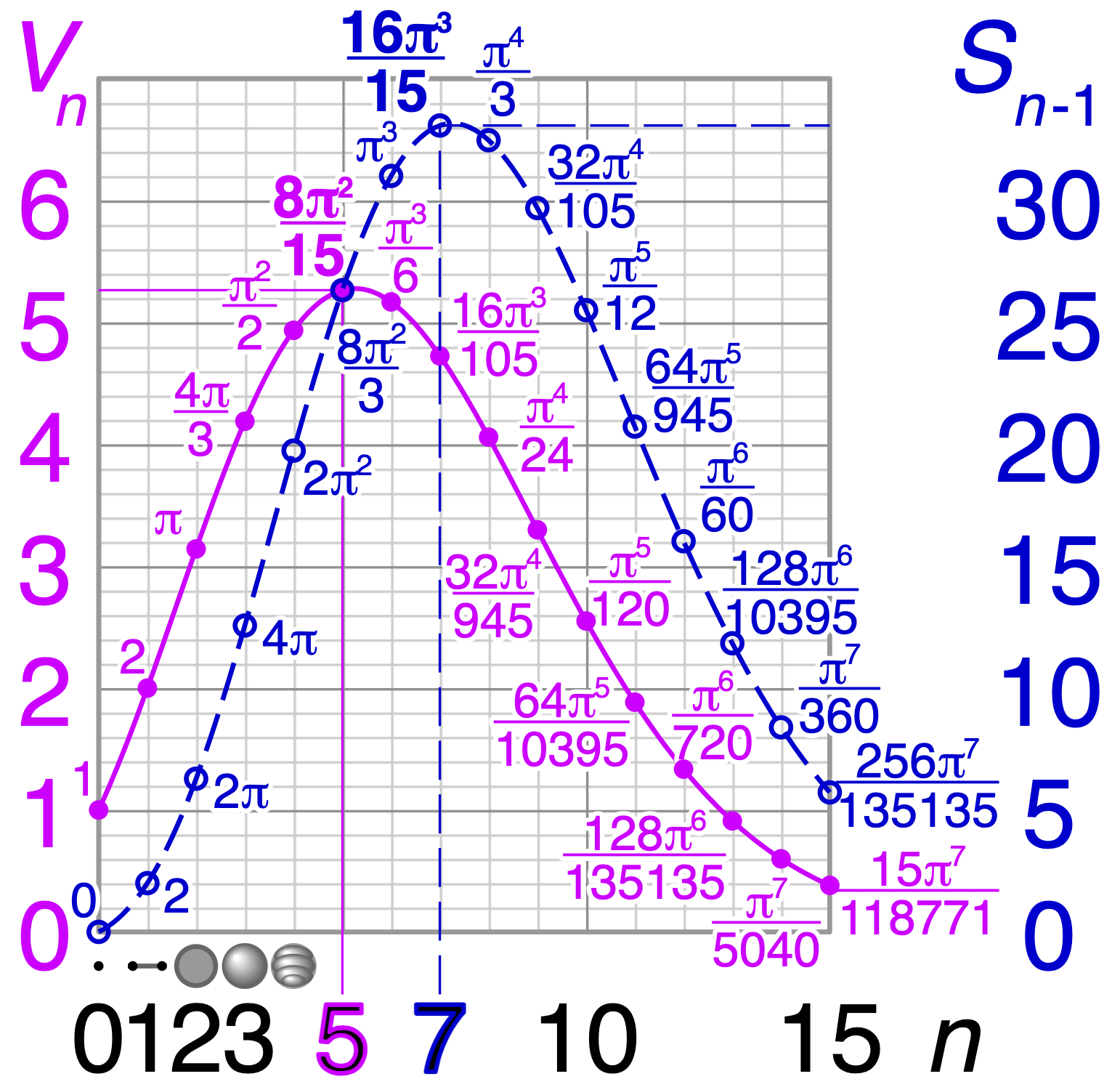}}
\caption{Volume and surface area of unit hypersphere vs dimensionality. \cite{wikipedia_image}}
\label{hypersphere}
\end{figure}

Consider the equations for a three dimensional unit sphere: \(x^2 + y^2 + z^2 = 1\), where the volume is simply \(4 \pi r^3 / 3\), which for a unit sphere is \(4\pi/3\). Now consider a hypersphere where we increase the number of dimensions governed by the similar formula \(w^2  + x^2  + y^2  + z^2  + … = 1\). To visualize in an abstract way, consider the difference between a perfect sphere and a collection of fronds at the top of a palm tree. As may be a surprising finding, for hyperspheres both the volume and surface area briefly increase with increasing dimensions until they reach a peak, after which point they progressively shrink to an asymptote at zero [Fig \ref{hypersphere}]. The paradox of hyperspheres is that with this decreasing volume and surface area, the expected distance between two sampled points will actually increase with parameterization \cite{Tu2002RandomDD}. Due to the curse of dimensionality, once a manifold starts to reach thresholds beyond order of 10 dimensions, evaluating fine-grained structure from random sampling becomes exponentially hard \cite{Intrinsic_Erba_2019}. Mathematicians currently have better understanding of hyperspheres in comparison to other high dimensional objects, even for simple shapes like hypercubes \cite{Granata_accurateestimation}, however this type of zero asymptotic volume convergence appears likely to arise when a shape is constrained through dimensional adjustment to a single scale, such as for a hypersphere could be the unit radius or for a loss manifold could be some loss value. We just don't know where the peak of the volume curve would occur. It is possible the machine learning community may have found another framing that can approximate the location of such a peak by way of the double descent interpolation threshold.

One way to think about the loss function of a neural network is as an unconstrained formula with weights (and other hyperparameters) as the variables, for example a loss function \(J\) may be derived as a function of weights \(J(w) = ?\) (which abbreviation is meant to generalize across loss metrics like mean absolute error or cross entropy), and through backpropagation we are trying to minimize \(J(w)\). However when you consider that a fitness landscape will in general have a global minimum, the loss function through backpropagation is shifted in direction towards a minimum loss \(L_{min}\) as \(J(w) \to L_{min}\). This also applies to any given value for \(L\), that is for any given loss, the formula \(J(w) = L\) is a constrained formula where each weight has some distribution of potential values associated with that loss, similar to how in a hypersphere there is some distribution of each variable associated with a specific radius. Thus \(J(w)\) can be approximated as a constrained formula around the weight set associated with the global minimum as well as for losses in the backpropagation states preceding reaching the global minimum, and where the volume of the distribution of weights are expected to contract as the loss approaches the global minimum as there become fewer weight sets capable of achieving better performance, and the volume will converge to a point (a single weight set) at the global minimum.

Gaps in loss manifold volume refer to the distinction for a training path of directional updates that result in increased loss from the prior epoch, which will not be considered a viable path barring some form of momentum or otherwise deviation from a pure gradient signal. The volume of a loss manifold can be considered in two framings: the volume of weight set distributions that are capable of realizing an exact loss value, or the volume of weight set distributions that are capable of achieving a loss lower than the prior epoch, in which case the second framing’s volume plus the volume of the corresponding gaps will equal the volume of the full range of initialization sampling. Note that an initialization sampled from a normal distribution means that there will be a (vanishingly small) probability of an infinite range of possible weight values sampled at initialization, thus when we talk about the volume of initialization sampling it needs a probabilistic element to be meaningful, e.g. volume of initialized weights for \(P_w > 10^{-10} \). 

A hypersphere aligned volumetric transience through increasing dimensions translated to our high dimensional loss function \(J(w)\) is really just another way of saying that with increasing dimensions by parameterization the degrees of freedom available to each weight corresponding to a given loss value will be diminished, kind of similar to what happens with L1 regularization which promotes collective sparsity of a weight set \cite{bengio2012practical}. However here we are not talking about the sparsity of a single collective weight set, more referring to sparsity of weight set distributions corresponding to a loss value (a loss manifold). This implies that individual weights will also result in, for a given weight \(w_i\), for that distribution of \(w_i\) corresponding to a given loss value, the sparsity of that single weight's distribution increasing with parameterization. With correlations, dimensionality's influence to individual weight distribution sparsity will be harder to see than sparsity across weights.

The main idea of regularization theory is to restrict the class of admissible solutions by introducing a priori constraints on possible solutions \cite{WebbFunctional}. Thus, with increasing dimensionality, a trend toward decreasing volume and surface area of hyperspheres could imply a corresponding trend towards increasing sparsity of each weight’s distribution associated with a loss value, which would enforce a kind of regularization by constraining degrees of freedom for weight sets traversed through a training path, explaining the inherent regularization of overparameterized networks. [Appendix \ref{app:other}] surveys related phenomena.

The preceding considers asymptotic dimensionality. The double descent phenomenon \cite{Belkin15849} may be associated with an additional correspondence to hypersphere volumes at lower dimensions where hyperspheres exhibit a peak in volume across dimensions [Fig \ref{hypersphere}]. Consider that the distribution of possible update steps available to an epoch after taking into account any stochasticity included in the training loop as a geometric figure with a dimensionality of its own. This differs from the geometric figure associated with the broader geometric regularization conjecture, which was simply the distribution volume of all weights associated with a loss value, as in this case we need to take account for the specific weight configuration at the point where a next epoch update step is considered for a training path, or more particularly the distribution of possible update steps available from each point, which distributions will have an intrinsic dimensionality \cite{https://doi.org/10.48550/arxiv.2001.10872} of their own varied along a training path tendril. We use the term ``tendril” to suggest that as the training path reaches loss values approaching the point of global minimum in the loss manifold, the surrounding intrinsic dimensionality of this smaller geometric figure will continue shrinking until reaching an effective zero dimensions at the global minimum, as in the ideal case at global minimum the loss manifold distribution at \(J(w) = L_{min}\), as a point, will have effectively zero dimensions. With correspondence to hyperspheres, at some point along that training path the intrinsic dimension of this distribution of possible gradient steps will retract across a peak in a volume to dimensions curve below which geometric regularization will re-emerge, visible when not masked by an adjacent regularizer, which explains the emergence of an epoch wise double descent.

A characteristic feature of double descent is the interpolation threshold for overparameterization which can be approximated as a minimum boundary where the number of parameters exceeds the number of training samples and below which threshold double descent does not manifest. We offer further speculative musings on origination in [\ref{apd:interpolationthreshold}].

It has been observed that the global minimum of an overparameterized model may not be a single point, it actually transitions at some scale of parameters to becoming a submanifold — as in having a range of possible weight sets all sharing a common loss value as the minima of the optimization’s fitness landscape \cite{weinan_ICML_keynote_2022}. We suspect that there is a very simple explanation that arises from the conjunction of the geometric regularization phenomenon coupled with the practicality of numeric representations of weights, activations, and gradients that collectively parameterize a loss function: it isn’t that geometric regularization is shrinking the count of available function representations, it is just that it is squishing them to numeric values falling below the capacity of the data type representing these parameters. (Not squished to underflow territory, more resulting in delta updates from gradient steps falling below the step size available to increment a value within the precision capacity of a data type representation’s bit registers).

The generalized Poincaré Conjecture, proven for dimensions \(>4\) \cite{10.2307/1970239} and later extended to 4 \cite{10.4310/jdg/1214437136} and 3 \cite{https://doi.org/10.48550/arxiv.math/0211159}, suggests that every simply connected, closed n-manifold is homeomorphic to the n-sphere, which demonstrates that topologically related manifolds have one-to-one point correspondence. Increased regularization with overparameterization could be interpreted as empirically suggestive of a similar dimensionally adjusted volumetric correspondence. This is somewhat of a conjecture, but with supporting evidence of well understood hypersphere geometry, extensive empirical demonstration, and the reality of a double descent found in both epoch wise and model capacity wise curves, we believe it is the most credible hypothesis to date.

\section{Related Work}\label{apd:relatedwork}

It has been known for some time that hyperspheres have asymptotic volumetric trends with increasing dimensionality \cite{doi:10.1080/0025570X.1989.11977419}. Some of our additional understandings of higher dimensioned geometries are less precise. Another channel of hypersphere theoretic research has been associated with finding optimal hypersphere packing densities at different dimensions, including notable recent results of proved optimal packing densities at 8 dimensions \cite{Viazovska_2017} and 24 dimensions \cite{Cohn_2017}. To our knowledge the primary published work from other authors seeking to relate properties of hyperspheres to neural network regularization was with respect to applying a constraint to weight updates in order to achieve hypersphere uniformity in a manner analogous to L2 regularization \cite{https://doi.org/10.48550/arxiv.2103.01649}, which is non-overlapping to our conjecture.

\begin{figure}
\centering
\centerline{\includegraphics[width=0.25\columnwidth]{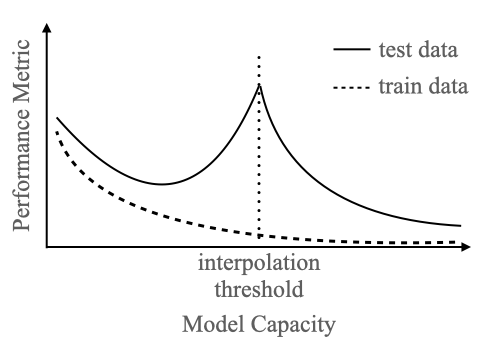}}
\caption{Model capacity double descent}
\label{doubledescent}
\end{figure}

Overparameterization is commonly considered as the set of neural network architectures with number of weights exceeding the complexity threshold where the count of weights equals that of training samples, although distinctions such as mild overparameterization verses heavy overparameterization may also come into play. A notable unexpected property of the overparameterization regime is that the conventional wisdom for the bias-variance tradeoff in training appears to be contradicted, with emergence of an ``epoch wise" double descent training curve in which progressing through epochs initially manifests overfit that reaches a peak before continued training results in a recovery of test performance to realize a better generalization than what was achieved prior to the overfit state \cite{https://doi.org/10.48550/arxiv.1912.02292}. The interpolation threshold beyond which the double descent phenomenon appears can be illustrated by charting performance curves for train and test data as a function of model complexity capacity [Figure \ref{doubledescent}] \cite{Belkin15849}, which exhibit what we call a ``model capacity" double descent, directly related to but distinct from the epoch wise double descent. Model capacity is often approximated by a ratio of weight to sample counts.

The overparameterization convention appears to have several benefits. Empirical studies demonstrate reduced risk of overfit \cite{NEURIPS2018_54fe976b} and remarkably small generalization error \cite{DBLP:conf/iclr/ZhangBHRV17}. The benefits appear to manifest across frameworks and modalities of application. Overparameterized models appear to result in smoother fitness landscapes with a smaller ratio of saddle points to global minima \cite{pmlr-v139-simsek21a}. The resulting models appear more robust to covariate shift, meaning distributional discrepancies between train and test data with retained label correlations \cite{Tripuraneni_overparam}, and their interpolations are smoother with a smaller Lipschitz constant \cite{bubeck2021a}. Although increasing parameters will have resource and latency impacts to inference, the resulting models can often be pruned with little or no cost to generalization \cite{Barsbey_heacytails}.

While there is often a material increase in time, cost, and complexity of training these models, the resulting performance characteristics appear to more than offset, and progressively higher thresholds of overparameterized transformers \cite{NIPS2017_3f5ee243} have led to natural language implementations with few shot learning capabilities like GPT-3 \cite{NEURIPS2020_1457c0d6} and several emerging foundation models since \cite{bommasani2021opportunities}. 

It has been somewhat of a mystery to researchers the source of this phenomenon. It has been demonstrated empirically that size alone does not explain it, and that some form of capacity control or implicit regularization is at play \cite{DBLP:journals/corr/NeyshaburTS14}. The phase transition to a double descent phenomenon has been directly linked to varying the ratio between number of parameters to samples in unregularized networks \cite{NEURIPS2020_37740d59}. There appears to be some relevance to model initialization consideration as models tend to learn a network close to the initialized random weights \cite{NEURIPS2018_54fe976b}. Counter to classical stochastic optimization theory, these models appear to train better with a constant SGD learning rate without momentum \cite{pmlr-v119-sankararaman20a}. Perhaps even more perplexing, aspects of the phenomenon are not limited to neural networks, with a similar double descent curve and generalization benefits with increasing model complexity capacity being demonstrated in other paradigms like kernel methods, nearest neighbors \cite{pmlr-v80-belkin18a}, decision tree paradigms like random forest and gradient boosting \cite{Belkin15849}, and quantum neural networks \cite{https://doi.org/10.48550/arxiv.2109.11676}.

\begin{figure}
\centering
\centerline{\includegraphics[width=0.25\columnwidth]{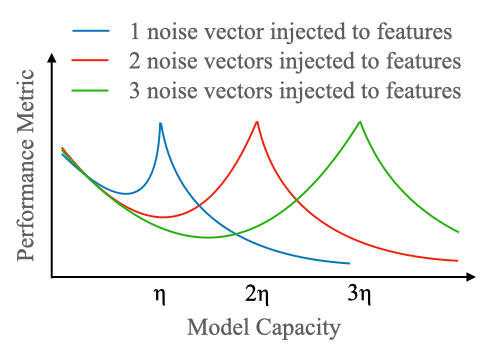}}
\caption{Perturbation vectors impact}
\label{perturbationvectors}
\end{figure}

Noise injections to training features result in an increased threshold for number of parameters needed to reach the overparameterization regime [Figure \ref{perturbationvectors}] \cite{pmlr-v139-dhifallah21a}, which we speculate is associated with additional perturbation vectors causing an increase to the intrinsic dimension of the modeled transformation function in a manner similar to data augmentation’s impact to intrinsic dimension \cite{marcu2021datacentric}.

The practice has mostly positive impacts to performance, although it does have the possibility of lazy training where a model converges exponentially fast to zero training loss to recover a linear model \cite{NEURIPS2019_ae614c55}, which may occur under some choices of hyperparameters. Properly sampled initializations benefit the optimization \cite{pmlr-v28-sutskever13} and help to avoid the undesirable lazy training phenomenon \cite{stoger2021small}. There have been reports of degradation of performance on under-represented subgroups, however this appears to resolve with better calibration to a classification output layer \cite{menon2021overparameterisation}.

Theoretical study of the phenomenon has followed several branches, this paper’s survey isn’t exhaustive. An influential channel of inquiry was to consider neural layers approaching the infinite width limit where the network's modeled function become a Gaussian distributed process at initialization, which assumption underlies the neural tangent kernel equivalency \cite{NEURIPS2018_5a4be1fa} that can be used to represent networks as a kernel function. This equivalent kernel’s positive definiteness \cite{fasshauer2011positive} can be used to evaluate network convergence properties, although this finding alone may not be sufficient to explain the impact of overparameterization since it appears to abate in presence of skip connections or batch normalization \cite{Goldblum2020Truth}. Other researchers have attempted to reason about translations in fitness landscape properties between different parameter regimes, which are closer aligned to the theme of this work. 

An overparameterized network is capable of learning any function represented by a corresponding network of fewer parameters \cite{Sun_SampleEffic}. This property appears to extend to networks of discrete activations learning the functions of smaller networks of smooth activations as has been proven for three layer networks modeling two layers \cite{NEURIPS2019_62dad6e2}. Networks have a universal connectivity property, so that if a modeled function may exist in parameter space it at least has the potential to be reached in training from a random initialization \cite{pmlr-v80-draxler18a}, which finding has also been extended to ReLU activations \cite{freeman2017topology}. It has been considered that a point in the fitness landscape of a network will translate to a corresponding manifold in the fitness landscape of a larger network \cite{pmlr-v139-simsek21a}, which observation is the closest we’ve seen in the literature to reasoning about geometry of loss manifold distributions as is the focus of this paper. An influential tangent line of inquiry has considered geometric properties of feature manifolds \cite{7974879}.

There appears also to be some differences in whether the source of overparameterization is from increasing network width or depth. It has been observed that wider networks are easier to train \cite{NEURIPS2018_5a4be1fa}, while deeper models have an implicit bias towards sparsity \cite{gissin2020the} and may exhibit loss manifolds with an increased prevalence of non-convexities \cite{NEURIPS2018_a41b3bb3}. Expected parameterization needed to reach generalization has been considered higher for deeper than wider networks, although it has recently been suggested that a mild overparameterization can also be used for deep networks \cite{chen2021how}. Deeper models have been shown to be more efficient at modeling higher complexity functions than shallow networks \cite{pmlr-v49-eldan16}. Gradient confusion refers to negatively correlated gradients between mini-batches, which has been shown to trend higher with deeper networks \cite{pmlr-v119-sankararaman20a}.

Implementing overparameterization in practice involves several considerations. Theorists have suggested choosing a width based on the point at which learning algorithms can provably learn a zero loss in non-convex training, and then if increasing the number of training samples the parameterization can be increased by widening layers in a corresponding manner \cite{Song_Subquadratic}. The duration of training may be balanced between scale of parameters and training tokens \cite{Zhai_2022_CVPR}. Deeper networks may realize similar benefits to wider networks with a common degree of mild overparmeterization \cite{chen2021how}. Discontinuous activations like ReLU are still appropriate and train faster than smooth activations \cite{Panigrahi2020Effect}. As noted above vanilla SGD with a constant learning rate has been found to outperform scheduled learning rate methods \cite{pmlr-v119-sankararaman20a}. The He initialization heuristic appears to lie at the boundary of the well-performing regime \cite{Song_Subquadratic}, which since the models tend to learn a model close to initialization is an important consideration. Note that overparmeterization can even be achieved by introducing intermediate linear layers that after training can be contracted algebraically to realize a more compact model \cite{NEURIPS2020_0e1ebad6}.

Recent work from \cite{https://doi.org/10.48550/arxiv.2203.15556} has demonstrated one can balance parameterization scale with number of training tokens in large language models for purposes of optimizing training-compute cost and performance. This is consistent with the geometric regularization conjecture as the asymptotic trend of a flattening volume curve may suggest a point of diminishing returns from overparameterization.

\section{Interpolation Threshold}\label{apd:interpolationthreshold}

Please consider the dialogue of this section as more of a speculative nature. We provided in [\ref{sec:hyperspheres}] a plausible conjecture for both the regularizing properties of overparameterized learning as well as a corresponding explanation for the epoch wise double descent phenomenon. Missing from the theory were considerations surrounding why the various phenomenon of overparameterized learning have been found to have a distinct boundary, known as the interpolation threshold, which is traditionally approximated as conditions where the number of model parameters exceed number of presented training samples \cite{Belkin15849}, manifesting a model capacity double descent [Fig \ref{doubledescent}].

We have already pointed out in [\ref{apd:relatedwork}] that the ratio of number of parameters to number of training samples alone may be insufficient to universally derive an interpolation threshold as feature properties of additional noise perturbation vectors also play a role [Fig \ref{perturbationvectors}]. It has also been demonstrated by \cite{Roberts_2022} that the dynamics of training a model is not just a function of parameter count but also of the architecture, e.g. considerations like depth and width. We offer in explanation that for two models of equal parameter count, the deeper model will have higher ratio of influence (ratio of parameter interactions / parameter count) between upstream and downstream weight updates during backpropagation, and that such channels of influence should contribute to increasing the complexity density of a neuron, and assuming a constant model complexity from parameter count this will translate to sparsity in weights, as such trends have been found for deeper architectures \cite{gissin2020the}. We also suggest that although the ratio of parameter count to training samples may have become a useful heuristic, a formal definition for the threshold would need to also account for intrinsic dimensionality of the data generating function, the complexity capacity of the model, and the complexity of the presented function at a given fidelity of representation. The dominant impact of number of training samples to the threshold should saturate if data is fully represented, on which we now attempt to extrapolate.

\begin{figure}
\centering
\centerline{\includegraphics[width=0.25\columnwidth]{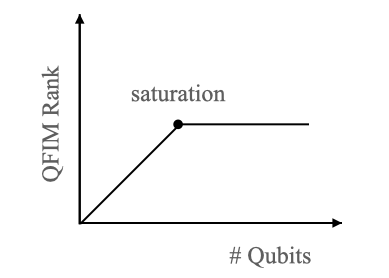}}
\caption{QFIM rank.}
\label{qfimrank}
\end{figure}

Let us note a few properties of a more recent paradigm of learning, that associated with quantum neural networks \cite{https://doi.org/10.48550/arxiv.2003.02989} comprised of parameterized quantum circuits \cite{https://doi.org/10.48550/arxiv.1802.06002}. Consider that a quantum sensor by definition will give our network access to the full superposition within the scope of input, at least prior to any measurement collapse. This differs from classical learning where the fidelity of a data generating function presented to a network may have representational gaps due to collected training data not covering the full surface of this manifold. (Or to put in plain English, the training samples may not capture all of the scenarios applicable to label generation.)


Various phenomenon of overparameterization are not limited to classical neural nets, as we noted in [\ref{apd:relatedwork}] they also arise both in legacy paradigms of learning as well as next generation quantum neural networks. However some recent work applicable to quantum neural networks have found results that appear unique to the quantum setting, for which we will shortly try to find parallels in classical networks. Quantum neural networks have a characteristic curve [Figure \ref{qfimrank}] \cite{https://doi.org/10.48550/arxiv.2109.11676} between model effective dimension (demonstrated by quantum fisher information matrix rank) and the number of parameterized qubits.


One may expect a similar curve for classical networks with the saturation point aligned to the interpolation threshold, however in the classical setting the spectrum of the Fisher information matrix is often more degenerate than quantum networks with more low magnitude eigenvalues, conditions that mainly arise in quantum networks with barren plateaus \cite{Abbas_2021}. We expect this is associated with different fidelities of received signals between quantum and classical learning. Before losing correlation at smallest scales, decreasing the number training samples in a classical corpus should present fidelities resembling a progression of the 2\textsuperscript{nd} law of thermodynamics from a disorder standpoint and a negative ``coffee progression" from a representational complexity standpoint; consider Appendix B.4 of \cite{Abbas_2021} in conjunction with Figure 2 of \cite{https://doi.org/10.48550/arxiv.1405.6903}.

\begin{figure}
\centering
\centerline{\includegraphics[width=0.8\columnwidth]{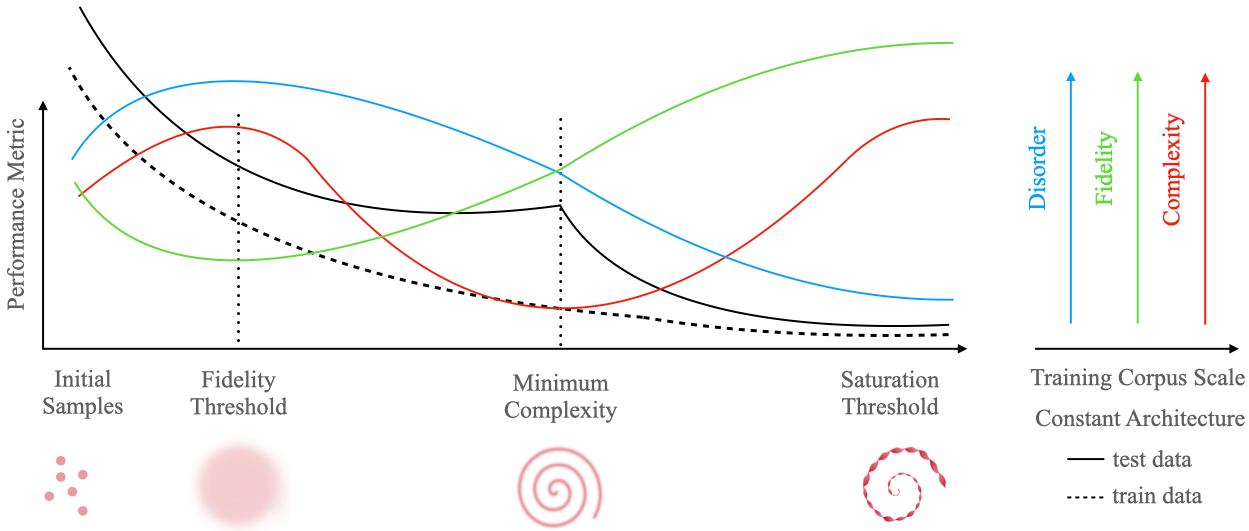}}
\caption{Data fidelity double descent.}
\label{fidelity}
\end{figure}

We hypothesize a data fidelity double descent curve based on varied scale of training data with fixed architecture [Fig \ref{fidelity}] which assumes overlapping convex or concave inflection points in fidelity and disorder. Precise features will likely be harder to manifest than a model capacity curve [Fig \ref{doubledescent}] because the contribution of removing a training sample from a corpus is reliant on composite distribution of other samples, so producing a uniform fidelity degradation would require either a sophisticated synthetic data convention or otherwise a scheduled sample retraction based on distribution in some manner. In practice the performance curves should demonstrate a stochastic progression, with peak overfit partly masked by the adjacent performance benefits of more training data preceding saturation. The initial complexity peak arises from spurious dimensions from gaps in fidelity of the training data. With a decrease in regularization at the interpolation peak model capacity that was applied to spurious dimensions will be diverted to overfit, and with re-emergence of geometric regularization at increasing data scales model capacity is applied to data complexity.


The concurrence of a model capacity and epoch wise double descent is suggestive to the universality of geometric regularization. Whether an interpolation threshold may be approximated by a ratio of parameter to sample counts or some combination of model capacity and data dimension at a given fidelity, it is arising from a different geometric form than the distribution of weights in a loss manifold. A remaining question is what may be a constraining constant through dimensional adjustment analogous to a hypersphere's radius or some loss function's loss value, as we expect a bottleneck is needed for n-sphere volumetric correspondence. We suggest this arises from a data representation's complexity at a presented fidelity in relation to a volume of diversity found in a model's distribution of representational forms, as the volume of a model's capacity for representational diversity will contract with overparameterization towards inherent generalization bias.

\section{Conclusion}
\label{sec:conclusion}

This paper has introduced the geometric regularization conjecture, which is associated with volume contraction with increasing parameterization of the distribution of possible weight sets associated with a loss value, and was inferred based on related properties demonstrated by hyperspheres. Geometric regularization would explain several phenomenon seen with overparameterized learning that have puzzled researchers. We believe double descent is a result of a training path reaching a loss value sufficiently close to a global minima that the distribution of possible weight set update steps corresponding to points reached in backpropagation will have a shrinking intrinsic dimensionality transience realized along the training path, resulting in this distribution retracting across its own peak in a volume verses dimensionality curve, after which there should arise a phase change as geometric regularization re-emerges when not masked by an adjacent regularizer. An interpolation threshold may need to account for complexity of data representation at different fidelities.


As many high dimensional properties are less tractable in today's theory, we expect the emergence of an empirical double descent curve with approximation methods for an associated interpolation threshold could become an alternate channel for mathematicians to investigate high dimensioned functions in future research. We explore additional weight distribution transience characteristics via loss manifold histograms in [Appendices \ref{app:histograms} - \ref{F}].


\bibliography{overparam_bib_deanon}
\bibliographystyle{icml2022}


\newpage

\appendix

\section{Table of Contents}
\label{A}

\begin{itemize}

\item \textbf{Appendix B}: \hyperref[app:other]{Other related phenomenon}
\item \textbf{Appendix C}: \hyperref[app:Higgs]{Saturated data}

\item \textbf{Appendix D}: \hyperref[app:histograms]{Histograms}
\item \textbf{Appendix E}: \hyperref[C]{ReLU activations}
\item \textbf{Appendix F}: \hyperref[D]{Width verses depth}
\item \textbf{Appendix G}: \hyperref[Eb]{Tail region}
\item \textbf{Appendix H}: \hyperref[E]{Initialization scaling}
\item \textbf{Appendix I}: \hyperref[F]{Additional training samples}
\item \textbf{Appendix J}: \hyperref[J]{Speculative speculations}

\end{itemize}

\newpage

\section{Other related phenomenon}
\label{app:other}

This appendix offers a brief speculative survey of how the geometric regularization conjecture might be related to other phenomenon observed with overparameterization. 

\begin{itemize}

\item \textbf{Retained proximity to initialization} \cite{NEURIPS2018_54fe976b}: The volume contraction of a weight distribution manifold in overparameterization if aligned to a hypersphere should converge at the infinite dimensional case to a projection in a one-dimensional world as a single point \cite{kuketayev2013probability}, explaining initialization proximity. It is a kind of paradox of the curse of dimensionality that with this convergence, the expected euclidean distance between two sampled points will increase with dimensionality.

\item \textbf{Smoother fitness landscape} \cite{pmlr-v139-simsek21a}: Borrowing the analogy that a point on a fitness landscape will translate to a manifold on a higher dimensional landscape [\ref{apd:relatedwork}], if the points on the higher dimensional fitness landscape are spread further apart \cite{Tu2002RandomDD}, then the landscape should appear to have a smoother characteristic when comparing comparably scaled weight deltas to an underparameterized model, which possibly explains the benefit of constant learning rates \cite{pmlr-v119-sankararaman20a}. The paradox of a point translating to a manifold of decreased volume also suggests some form of contraction from a manifold of similar points on an underparameterized landscape into aggregated smaller volume manifold with overparmeterization. The observation of reduced concentration of saddle points in an overparameterized fitness landscape \cite{pmlr-v139-simsek21a} suggests these have a higher prevalence of consolidations. 

\item \textbf{Regularization dampening double descent} \cite{Belkin15849}: We suspect the result that regularization dampens the interpolation peak at a minimum is associated with a more consistent degree of regularization through training, as geometric regularization may not be a dominant feature until the training path reaches well into the left tail of the loss manifold distributions. Without a secondary regularization at play a training path in effect will experience a phase transition from no regularization to dominant geometric regularization near the interpolation peak.

\item \textbf{Dropout regularization} \cite{JMLR:v15:srivastava14a}: Our discussions related to wide vs deep networks may be relevant to dropout regularization. Consider that when randomly dropping neurons, the network is then channeling backpropagation in a path consistent with what would be realized for a narrower width network with an increased ratio of influence, with implications for sparser representations.

\item \textbf{Other learning paradigms} \cite{pmlr-v80-belkin18a}: Our comparison of the hypersphere to a set of neural network weight distributions associated with a loss value we expect will be equally valid for any learning paradigm in which a large number of parameters are tuned in the direction of a decreasing loss signal toward a global minimum. In any form, the volume of the distribution of weight configurations should constrict with increased dimensionality, and an epoch wise double descent may manifest when weights traverse a tendril of shrinking effective dimensions.

\item \textbf{Gradient confusion} \cite{pmlr-v119-sankararaman20a}: With a thicker left tail in the histogram space [\ref{app:histograms}] as well as a tendency for reduced sparsity, the set of weight configurations that can approximate a modeled function will be larger for wider networks, and the corollary is that deeper networks will have a greater density of gaps in the weight distribution manifold associated with loss values exceeding that realized in the prior epoch, such that the statistical variations between mini-batches may result in diverting the direction of a training path due to proximity to one of these gaps, causing increased gradient confusion for deeper networks.

\end{itemize}

\newpage
\section{Saturated data}
\label{app:Higgs}

For empirical evaluation of the interpolation threshold without dominant influence of sample count we expect a data set represented in higher fidelities would help. We offer that for the classical setting, the tabular modality may have a unique potential to have data generating functions presented to a network by a training corpus in a fully represented manner at reasonable scales. Compare to the image modality, where features have a near infinite range of light sources, rotation angles, camera angles, or object compositions. Fully representing such complexity through a training corpus may take orders of magnitude more samples than tabular, with a model's complexity capacity otherwise diverted into spurious dimensions.


\begin{table}[h]
\caption{Higgs Benchmarking (AUC)}
\label{sample-table}
\begin{center}
\begin{tabular}{lll}
\multicolumn{1}{c}{\thead{\bf  }}  &\multicolumn{1}{c}{\thead{\bf Raw Data }}  &\multicolumn{1}{c}{\thead{\bf Noise Augment \\   }}  
\\ 
{\thead{\bf full data \\ 42 epochs}}    & \thead{ 0.8670 \\ -} & \thead{  0.8667 \\\bf  (0.0003) } \\
\hline
{\thead{\bf 5\% data \\ 14 epochs}}    & \thead{ 0.8439 \\ -} & \thead{  0.8453 \\\bf  0.0014} \\
\hline
{\thead{\bf 0.25\% data \\ 3 epochs}}  & \thead{ 0.7718 \\ -}   & \thead{  0.7821 \\\bf  0.0103} \\
\end{tabular}
\end{center}
\end{table}

As what may be considered at least suggestive evidence to demonstrate saturation of a data set representation by a training corpus, consider the tabular data augmentation by noise injection benchmarking applied to the Higgs data set \cite{Baldi:14} with a fastai learner \cite{Howard_2020} detailed in \cite{https://doi.org/10.48550/arxiv.2202.09496}. The authors found that when presenting the entire training corpus to a neural network, the benefit of data augmentation by noise injection was negligible. However when the training corpus was pared down to decreasing scale of samples, the benefits of data augmentation by noise injection appeared to grow proportionally to reductions in scale. This suggests that the Higgs data set may be a useful resource for researchers seeking to consider a fully represented classical training corpus in further inquiries on this matter.

\newpage
\section{Histograms}
\label{app:histograms}

We sought to visualize loss manifold geometries by exploring meta properties of weight set distributions with a kind of monte carlo evaluation to derive histograms of binary cross-entropy loss values realized from randomly initialized weight sets. We haven't seen loss manifold histograms considered in this manner by the literature, where varied setups were used to explore patterns that became apparent at the smallest scales. We drew some inspiration from Stephen Wolfram’s explorations of cellular automata and hypergraphs, in which he surveyed and catalogued various patterns that arose across simulated configurations. Such exploration is the only intent of these appendices, as we found the exercise helped us gain intuitions. We consider most of these explorations as tangential to the scope of the paper rather than directly supportive of the main dialogue.

The histograms were prepared in a series of jupyter notebooks on the Colaboratory platform. A representative notebook is provided with the supplemental material, where each of the setups had some degree of variation over this template. The network architecture was initially modeled as formulas in the cells of a spreadsheet, we soon found we could incorporate more elaborate architecture conventions and larger depth with the support of Tensorflow \cite{tensorflow2015-whitepaper}. We used samples from the Titanic data set as features. The notebook effectively demonstrates the initialization of a small network based on specification of width, depth, activation, and initialization type. Once a network was initialized with sampled weights, the features were passed to a predict operation similar to what would be performed on a trained model in inference, however in this case the predict was applied on the initialized weights without any form of training. The output of the predict and the corresponding labels were then fed to a binary cross entropy loss evaluation (without the ``from\_logits" option) to realize a single loss value recorded as one count within the histogram. Histogram aggregation involved repeating this setup a number of times based on either a designated sample count or in some cases running samples for duration of the Colaboratory 24 hour run time window.

We limited the feature count to three, with one numeric and two categoric, because we expected that the more features that were applied, the larger the scale of sampling would be required to get reasonable representation in histogram tail regions. For similar reasons, we focussed most of our inspections on loss values calculated for a single sample. We acknowledge this is borderline trivial territory with only three features, we found that even in this simplest of setups there were still characteristic patterns that emerged that we hope may be further investigated at more extensive scales by us or others with additional resources in future work. 



Although these histograms didn't reveal the distribution volume itself, they do demonstrate the relative volumes between different loss values for a given network configuration as a loss with higher number of possible weight configurations should stochastically demonstrate an increased probability of representation from random sampling, with improved fidelity from an increased number of samples. By comparing trends across histograms in varied network configurations, we hoped to infer aspects of distribution property effects realized from different degrees of overparameterization, width, depth, activation functions, initializations, and etc. At the scale of sampling we applied these histograms often didn't include representation of weight configurations associated with global minimum, however in many cases, and especially when considering loss from a single training sample, there were characteristic shapes and trends demonstrated in the left tail of the distribution which is the area of most interest for considerations of double descent.

\begin{figure}[ht]
\vskip 0.2in
\begin{center}
\centerline{\includegraphics[width=0.6\columnwidth]{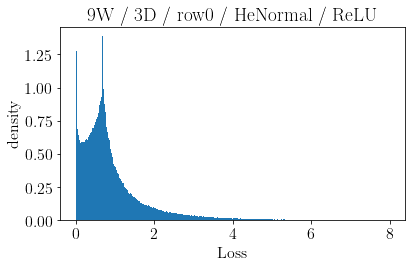}}
\caption{One training sample, dominant central mode}
\label{histogram_example_1}
\end{center}
\vskip -0.2in
\end{figure}

A recurring characteristic was the presence of a central mode in the distribution (visible as a peak), which appeared to universally align with the loss value realized from a 0.5 sigmoid output, it turned out the peak was a relic from the use of ReLU activations which at such small widths may often return all zero values in a preceding layer. Subsequent experiments with an increased number of features demonstrated that the number of sampled peaks appeared to correlate with number of features. Although some of the peaks phased in and out with small variations of parameters and configurations in a manner resembling the appearance of harmonics, we suspect the effect was from bin boundaries established by sampled minimum, and in aggregate trends were still visible. 

When we ran comparable setups with tanh for a smooth activation, the mode transitioned from one or more peaks to a single moded curve. In many cases for ReLU activations a second mode would appear. We speculate that the zero value secondary mode, as visible in Figure \ref{histogram_example_1}, might be associated with the phenomenon of lazy training \cite{NEURIPS2019_ae614c55} noted above. In a small number of cases a second mode also appeared in the right tail.

\begin{figure}[ht]
\vskip 0.2in
\begin{center}
\centerline{\includegraphics[width=0.6\columnwidth]{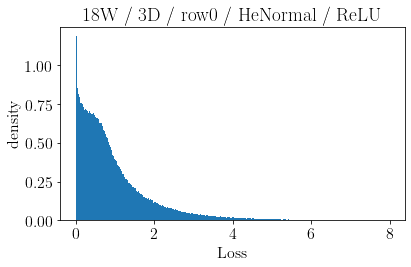}}
\caption{One training sample, dominant zero mode}
\label{histogram_example_2}
\end{center}
\vskip -0.2in
\end{figure}

We focussed most of our attention on histograms derived from a single sample to promote trend visibility. Two characteristic patterns appeared, one with the central mode and a reduced left tail [Fig \ref{histogram_example_1}], and the second with progressive volume towards a dominant zero mode [Fig \ref{histogram_example_2}]. In some cases these two conventions appeared with the same network configuration when evaluated against different training samples without He scaling. We expect these two cases may be aligned with what has been shown by other researchers when training on a single sample of having two scenarios of a model either memorizing or learning a representation \cite{DBLP:conf/iclr/ZhangBHRV17}. 

\begin{figure}[ht]
\vskip 0.2in
\begin{center}
\centerline{\includegraphics[width=0.6\columnwidth]{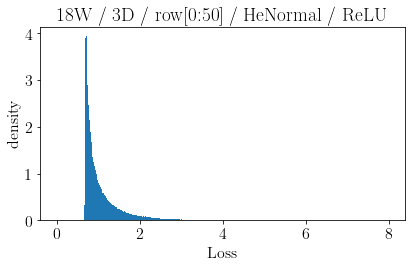}}
\caption{50 training samples, invisible left tail}
\label{histogram_example_50samples}
\end{center}
\vskip -0.2in
\end{figure}

The distinction between evaluations of a single training sample verses multiple training samples was notable. When loss was averaged over multiple training samples the left tail representation (for values below the primary mode) was greatly diminished, in most cases not visible at our sampling rate [Fig \ref{histogram_example_50samples}], although we could still confirm the tail existed for a given architecture by training the model for a few epochs, which would realize a loss value below the minimum sampled in the histogram. This is consistent with expectations that a weight set that can represent transformations of multiple samples is much rarer than one representing a single sample. That this kind of central mode dominated distribution would arise even when aggregating loss across samples with majority zero mode dominant distributions like Figure \ref{histogram_example_2} suggests that the zero mode dominant distributions for each sample have weight set distributions that are mostly non-overlapping.

We did not find that the aggregate histograms strongly aligned with any of the traditional left tail bounded distributions like lognormal, gamma, or Weibull (we evaluated a few with statistical tests using the Wolfram Language \cite{WolframLanguage} by deriving distribution parameters with FindDistributionParameters and then deriving a p-value with DistributionFitTest). However they still demonstrated some characteristic features of single mode distributions, and after averaging across multiple samples any secondary modes appeared to contract towards the central until losing visibility [Appendix \ref{F}]. Among those features was the presence of a single mode, what appeared to be an unbounded right tail, and a bounded left tail. Again this left tail in many cases became invisible to our sampling rate, however we could infer from assessing a loss value after inference from a comparable architecture trained to overfit and the connectivity principle \cite{pmlr-v80-draxler18a} that such a tail exists.

One way to think about what is taking place with the histograms is that we are compressing the geometry of loss manifold distributions down to a binary cross entropy comparison between sampled state and a designated lowest energy state, where lowest energy refers to the case where the sampled state's transformation function matches the natural label generating function at a global minima \(L_{min}\). Note that this lowest energy state is not an inherent property of the geometry, it is low energy in comparison to some targeted label generating function likely of different intrinsic dimensions than the capacity of the weight set. The sampled loss density is a kind of proxy for volume, and we can infer by the shape of the histogram curve which loss values will have larger manifold volumes relative to other loss values from the same architecture and initialization, as well as geometry transitions (i.e. surrounding volume expansion or contraction) that will be seen by an optimizer along a training path.

\twocolumn

\section{ReLU activations}
\label{C}

\begin{figure}[ht]
\vskip 0.2in
\begin{center}
\centerline{\includegraphics[width=0.7\columnwidth]{Fig2_010922_.png}}

\centerline{\includegraphics[width=0.7\columnwidth]{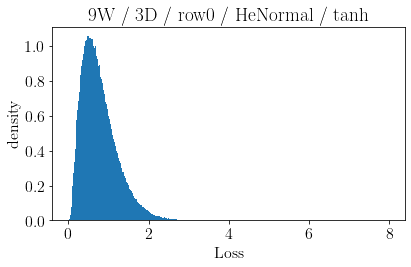}}
\caption{Corresponding ReLU (top) and tanh (bottom) example 9 wide}
\label{CorrespondingReLUtanh_1}
\end{center}
\vskip -0.2in
\end{figure}

\begin{figure}[ht]
\vskip 0.2in
\begin{center}
\centerline{\includegraphics[width=0.7\columnwidth]{Fig3_010922_.png}}

\centerline{\includegraphics[width=0.7\columnwidth]{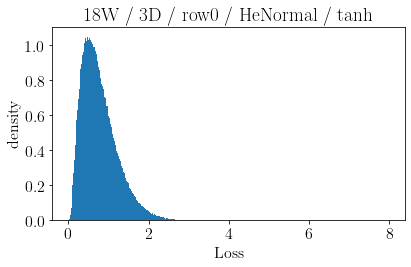}}

\centerline{\includegraphics[width=0.7\columnwidth]{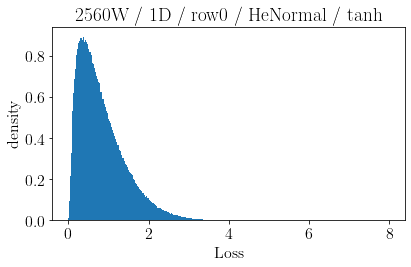}}

\caption{Increased width to 18, ReLU (top) and tanh (middle) // tanh approaching infinite width (bottom)}
\label{CorrespondingReLUtanh_2}
\end{center}
\vskip -0.2in
\end{figure}

The distinction between ReLU \cite{nair2010rectified} and tanh was noticeable [Fig \ref{CorrespondingReLUtanh_1}]. Relu exhibited sharp peaks while tanh was more of a smooth curve with only one visible mode. Tanh was more stable with architecture variations. In some cases Relu would shift noticeably while the corresponding tanh would be hard to distinguish [Fig \ref{CorrespondingReLUtanh_2}]. However when we modeled architectures approaching infinite width, and especially with shallower networks, a slight zero mode shift could be seen in the tanh [Fig \ref{CorrespondingReLUtanh_2}]. The higher density of low loss scenarios with ReLU offers an explanation for empirical benefits whose mechanism has been somewhat of a mystery \cite{Roberts_2022}. We expect the cause of the higher density is that by making a zero activation a range instead of a point, it increases the distributional density of narrower width models in the loss manifold classical superposition, increasing the diversity of representational capacity available to a model. The discontinuity in the activation in lieu of continuous curve allows for increased diversity of adjacent representational forms available to an epoch.

\section{Width verses depth}
\label{D}


Deviations of width and depth produced inverse directions in transitions from regimes similar to Figure \ref{histogram_example_1} and Figure \ref{histogram_example_2}. Histograms with the dominant central mode shifted density in direction of the dominant zero mode with increasing width, and more depth shifted in the counter direction of increased density to the central mode. The trends endured with inverted labels or batch normalization.

Consider network architectures approaching infinite width. The neural tangent kernel framing \cite{NEURIPS2018_5a4be1fa} suggests that these will converge to weights modeling a Gaussian distributed process at initialization. This known property of the modeled function may become an interesting channel for researchers to relate histogram properties to properties of a resulting function. When we modeled large width scenarios in comparison to corresponding parameterized deeper models with a tanh activation, which didn't exhibit the absolute zero mode like ReLU, we found the dominant mode also trended leftward when approaching asymptotic width. 

\begin{figure}[ht]
\vskip 0.2in
\begin{center}
\centerline{\includegraphics[width=0.7\columnwidth]{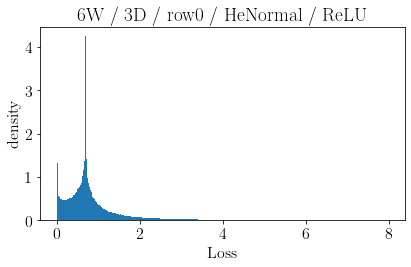}}

\centerline{\includegraphics[width=0.7\columnwidth]{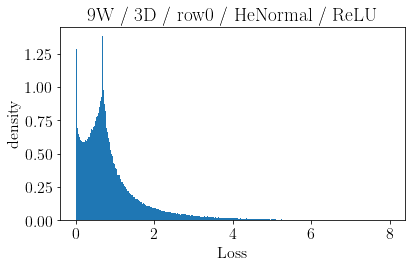}}

\centerline{\includegraphics[width=0.7\columnwidth]{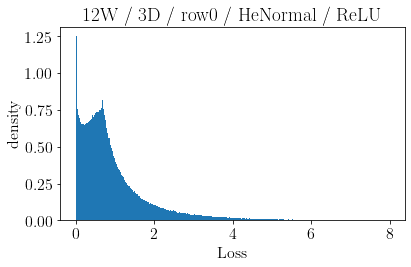}}
\caption{Transitions through increasing width}
\label{WideningLayersExample}
\end{center}
\vskip -0.2in
\end{figure}

\begin{figure}[ht]
\vskip 0.2in
\begin{center}
\centerline{\includegraphics[width=0.7\columnwidth]{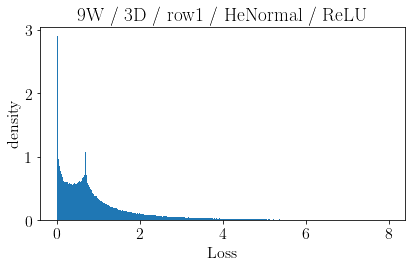}}

\centerline{\includegraphics[width=0.7\columnwidth]{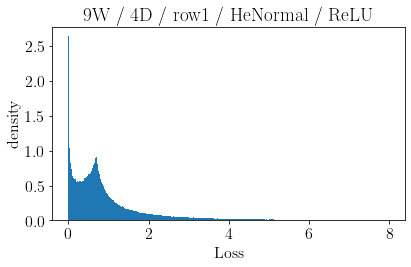}}

\centerline{\includegraphics[width=0.7\columnwidth]{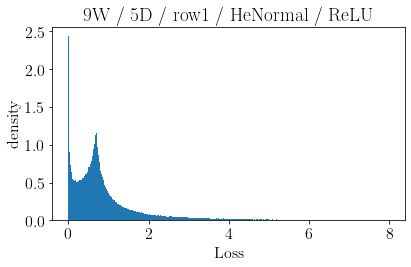}}
\caption{Transitions through increasing depth}
\label{DeepeingLayersExample}
\end{center}
\vskip -0.2in
\end{figure}

This appendix offers a few representative examples  implications of wide verses deep networks. [Fig \ref{WideningLayersExample}] demonstrates the transition from increasing width of a three layer network through 6, 9, and 12 neurons, demonstrating the characteristic shift towards a dominant zero mode. [Fig \ref{DeepeingLayersExample}] demonstrates a similar transition through increasing depth.

\section{Tail region}
\label{Eb}

It is probably worth reiterating that most of these discussions so far have focused on histograms derived from single training samples. As additional samples are added to the inference basis of the loss values, the left tail of the histogram distribution quickly shrinks to invisibility for our extent of sampling which aligns with intuition. Even though we don't have visibility of this tiny left tail, we can infer properties from what we have demonstrated takes place in loss manifold distributions on single training samples, after all the aggregate binary cross entropy loss value is derived from the average of the loss from each training sample. Thus if we can identify trends in the histograms of single training samples, we can expect they will also manifest in the invisible left tail of the aggregate histogram across all training samples. 

\begin{figure}[ht]
\vskip 0.2in
\begin{center}
\centerline{\includegraphics[width=0.7\columnwidth]{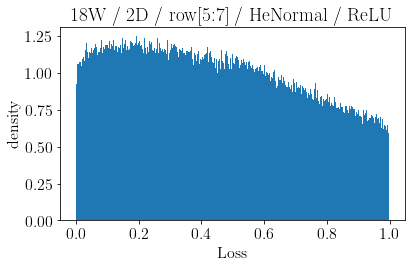}}

\centerline{\includegraphics[width=0.7\columnwidth]{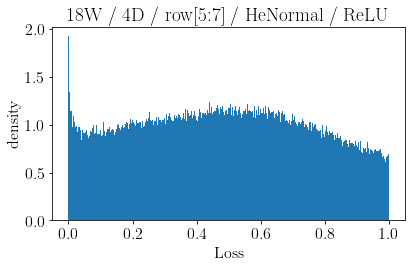}}

\centerline{\includegraphics[width=0.7\columnwidth]{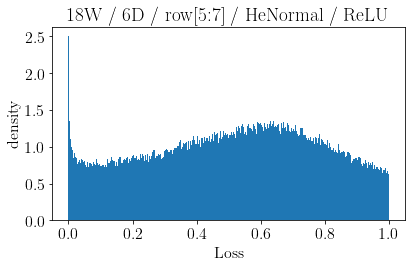}}
\caption{Tail region closeup with depths 2-6}
\label{Tail_Examples1}
\end{center}
\vskip -0.2in
\end{figure}

\begin{figure}[ht]
\vskip 0.2in
\begin{center}
\centerline{\includegraphics[width=0.7\columnwidth]{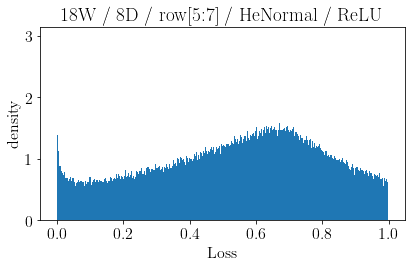}}

\centerline{\includegraphics[width=0.7\columnwidth]{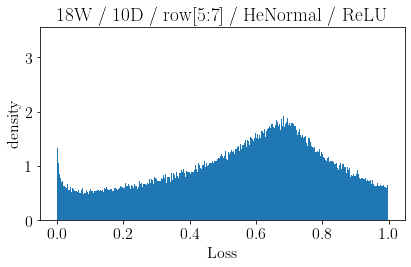}}

\centerline{\includegraphics[width=0.7\columnwidth]{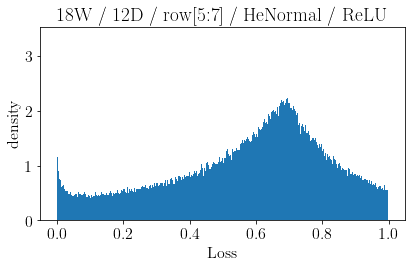}}
\caption{Tail region closeup with depths 8-12}
\label{Tail_Examples2}
\end{center}
\vskip -0.2in
\end{figure}

We found from the depth and width experiments that wider networks will have a greater proportion of low loss value weight configurations available than equivalently parameterized deep networks, which also aligns with the ratio of influence framing [\ref{apd:interpolationthreshold}]. What we didn't know is what the aggregate loss manifold volume was in comparisons between wider and deeper networks at a common loss value. Just because a deeper network has a stronger representation in the central mode region of sampled loss values, it might still have similar volumes of loss manifolds in the left tail in comparison to wider networks. The histograms only reveal relative distribution volume relationships associated with different loss values for a common architecture within the same sampling operation. We attempted to circumvent this challenge by continued sampling of single sample configurations until reaching a common threshold for number of sampled left region values between configurations in order to evaluate the region in isolation. This approach yielded a similar pattern, with increasing depth causing a transition from zero mode to central mode dominated characteristics [Fig \ref{Tail_Examples1}, \ref{Tail_Examples2}].

This finding suggests that for aggregated loss values across multiple training samples, where the left tail region becomes invisible to our sampling assessment and these zero mode verses central dominated profiles average out across samples to a central mode profile, the path followed by backpropagation towards minimum loss will traverse through a different profile of loss manifold geometry transitions in wider verses deeper networks. We expect the left tail region of wider networks will thus have thicker tails in this histogram space than corresponding deeper networks, suggesting that the geometric regularization experienced by deeper networks will be of greater intensity once reaching sufficient depth into the left tail assuming that an increased number of manifold gaps doesn't obstruct the path. We expect that the convention that thicker tails correspond to increased kurtosis may not apply for this consideration since the emergence of tail thickness is associated with presence of a second mode in the aggregated loss values as opposed to a dispersion from central mode. The idea that deeper networks will have greater constraints on weight configurations is also consistent with deeper networks trending towards more sparse weight configurations as seen by \cite{gissin2020the}.

We thus believe that wide and deep architectures may offer a tradeoff between number of minima and complexity capacity of the network as evidenced by zero mode dominance and trends toward sparsity.


\section{Initialization scaling}
\label{E}

Properly scaled initialization benefits optimization \cite{pmlr-v28-sutskever13}, and He initialization \cite{he2015delving}, which scales initialized sampling based on architecture of dimensions of inputs to a layer, appears to lie at the boundary of the well-performing regime \cite{Song_Subquadratic}. 

\begin{figure}[ht]
\vskip 0.2in
\begin{center}
\centerline{\includegraphics[width=0.7\columnwidth]{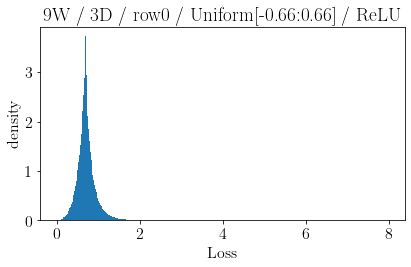}}

\centerline{\includegraphics[width=0.7\columnwidth]{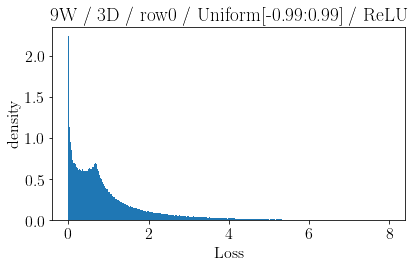}}
\caption{Random Uniform initialization with increasing scale}
\label{Initialization_towards_zero_mode}
\end{center}
\vskip -0.2in
\end{figure}

Note that He differs from Xavier initialization by not including the layer output dimensions in the scale derivation's denominator, resulting in a larger scale, and has been demonstrated as more suitable for nonlinear activations like ReLU \cite{kumar2017weight}. He initialization samples from Gaussian, or a similar scaling may be adapted for sampling from a Uniform distribution. One of our experiments involved aggregating histograms with Normal and Uniform, and then comparing each again to other scales. Interestingly, He scaled Normal [Fig \ref{histogram_example_1}] and Uniform shared a similar appearance of histogram characteristics, which for the demonstration architecture exhibited a balance between central and zero mode, however when we increased the scale of Normal and compared to a corresponding Uniform variation, it appeared that Uniform was more quick to shift into a dominant zero mode [Fig \ref{Initialization_towards_zero_mode}], suggesting that Normal is more stable than Uniform. 

We interpret the alignment of increased initialization scale of shifting the histogram towards dominant zero mode having a similar effect to widening layers as suggesting that networks have a larger proportion of low loss weight configurations when the input to layers have greater range of activations. The introduction of batch normalization did not appear to change this property, confirming that it was not arising from weight magnitude but spread. Increasing initialization scale is known to help break symmetry between units \cite{GoodBengCour16} and impact the implicit regularization of gradient descent \cite{Ba2020Generalization}. The paradox of increased regularization with scale \cite{pmlr-v75-li18a} in conjunction with a zero mode dominated tail possibly suggests for wider networks, which also approach a zero mode, the minima may be more spread out in the loss manifold than for deeper networks. One might expect that such mode balance could be used to align width and depth configurations when adding more training samples, however this has been demonstrated as less influential than simple parameter count \cite{https://doi.org/10.48550/arxiv.2001.08361}. This implies that width and depth configuration [\ref{apd:interpolationthreshold}] is more influential to the grouping characteristics of minima in the fitness landscape than to model complexity.

\begin{figure}[ht]
\vskip 0.2in
\begin{center}
\centerline{\includegraphics[width=0.7\columnwidth]{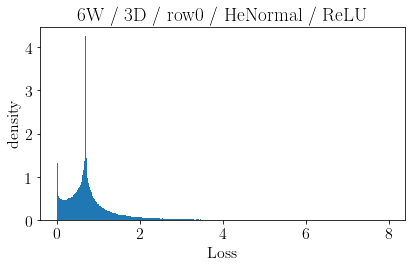}}

\centerline{\includegraphics[width=0.7\columnwidth]{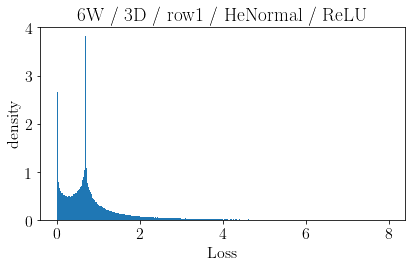}}

\centerline{\includegraphics[width=0.7\columnwidth]{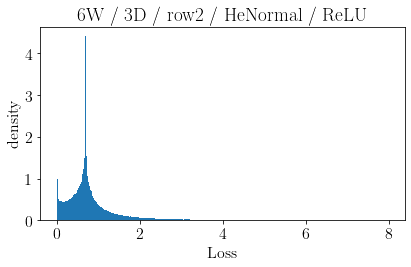}}

\caption{He initialization with different training samples}
\label{Hevssamples}
\end{center}
\vskip -0.2in
\end{figure}

\begin{figure}[ht]
\vskip 0.2in
\begin{center}
\centerline{\includegraphics[width=0.7\columnwidth]{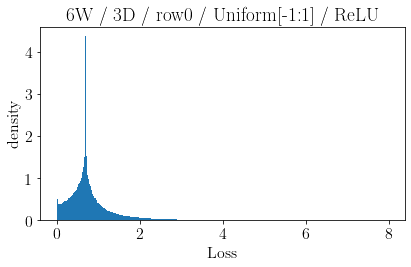}}

\centerline{\includegraphics[width=0.7\columnwidth]{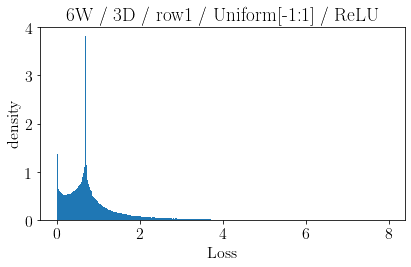}}

\centerline{\includegraphics[width=0.7\columnwidth]{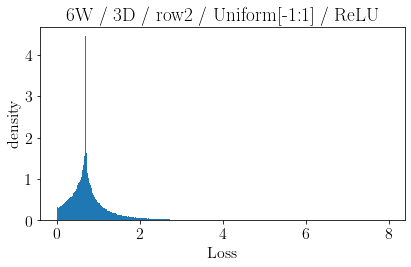}}

\caption{Uniform initialization with different training samples}
\label{uniformvssamples}
\end{center}
\vskip -0.2in
\end{figure}

We noted prior that in some cases dominance between the two mode conventions appeared with the same network configuration when evaluated against different training samples [\ref{app:histograms}]. It turns out this appeared to be more prevalent when applying initializations sampled from a uniform instead of normal distribution, as demonstrated here He scaled Normal appears to be stable across these three representative training samples [Fig \ref{Hevssamples}], while Uniform, with an arbitrary scale of +/-1, has some deviation in mode dominance characteristics across those same samples [Fig \ref{uniformvssamples}]. 

\section{Additional training samples}
\label{F}

\begin{figure}[ht]
\vskip 0.2in
\begin{center}
\centerline{\includegraphics[width=0.7\columnwidth]{Fig4_010922_.png}}

\centerline{\includegraphics[width=0.7\columnwidth]{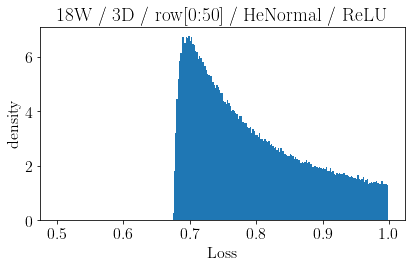}}
\caption{Close-up of 50 training samples}
\label{histogram_example_50samples_closeup}
\end{center}
\vskip -0.2in
\end{figure}
\begin{figure}[ht]
\vskip 0.2in
\begin{center}
\centerline{\includegraphics[width=0.7\columnwidth]{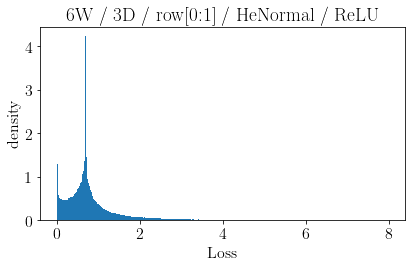}}

\centerline{\includegraphics[width=0.7\columnwidth]{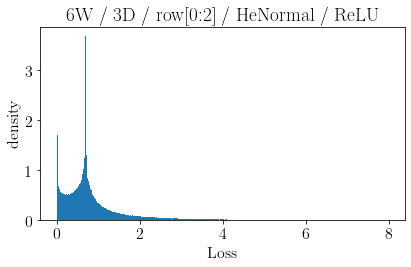}}

\centerline{\includegraphics[width=0.7\columnwidth]{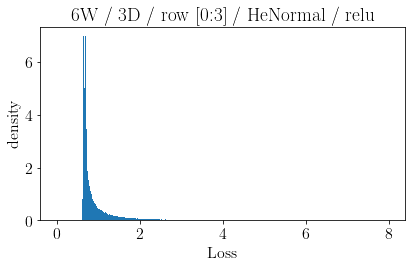}}

\caption{Scenarios of 1, 2, and 3 training samples}
\label{addingsamples}
\end{center}
\vskip -0.2in
\end{figure}
\begin{figure}[ht]
\vskip 0.2in
\begin{center}

\centerline{\includegraphics[width=0.7\columnwidth]{Notebook78_add_samples3.png}}

\centerline{\includegraphics[width=0.7\columnwidth]{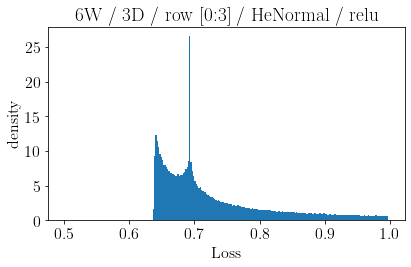}}

\caption{Close-up of 3 training samples}
\label{threesamples_closeup}
\end{center}
\vskip -0.2in
\end{figure}

Most of the histograms in these appendices have inspected a single training sample for tail visibility. Evaluating 50 concurrent training samples produced significant tail contraction [Fig \ref{histogram_example_50samples_closeup}]. To help isolate a transition point, in [Fig \ref{addingsamples}] a histogram is shown based on 1, 2, and 3 aggregated samples. It appears that the first two samples were similar enough that no complicated function was needed to relate their inference, so the left tail was retained with their aggregation. The addition of a third sample resulted in the left tail compressing to close proximity to the central mode [Fig \ref{threesamples_closeup}]. Note that a small indication of a left tail is visible adjacent to the leftmost peak. 

\newpage\
\onecolumn

\section{Speculative speculations}
\label{J}

We hope that the reader may humor us with these few additional, more speculative, speculations about the implications of the conjectures in this work.

\begin{itemize}

\item \textbf{Conservation of complexity}: The interpolation peak in the data fidelity double descent curve [Fig \ref{fidelity}], as well as what we have noted about expectation that increasing the ``ratio of influence" for increased neuron information density is more impactful to minima grouping characteristics in the loss manifold than it is to the model complexity [Appendix \ref{E}], suggests to us that there may be a form of complexity conservation for a given architecture at a given loss value, at least from a classical superposition standpoint, although we expect if you consider the distribution of models in a quantum neural network there may be equivalent performing models in classical space but with different complexities in quantum space, similar to what we have seen referred to as partial equivalence \cite{https://doi.org/10.48550/arxiv.2208.07564}.

\item \textbf{Barren plateaus}: We noted in [\ref{apd:interpolationthreshold}] that a degenerate quantum fisher information matrix rank resembling the low fidelity scenario for classical networks is primarily seen with quantum neural networks in cases of barren plateaus \cite{Holmes_2021}. We speculate that there may be a relevance to the complexity and disorder progression noted in [Fig \ref{fidelity}], such that as disorder progresses with increasing entropy for a system by the 2\textsuperscript{nd} law, it will cause some superposition elements to be inaccessible to a trained model. We expect that these obscured regions may manifest to a quantum network's loss manifold as a barren plateau.

\item \textbf{Generalization bias}: We used the term ``inherent generalization bias" at the close of [\ref{apd:interpolationthreshold}]. This is a somewhat remarkable concept, and we think that there may be relevance to theoretical physics. Consider what \cite{Roberts_2022} noted that we can actually fully train infinite-width networks in one theoretical gradient-descent step. That is, we can take a giant leap right to the minimum of the loss. (pg 252), which can be achieved independent of the applied algorithm (pg 257). \cite{Roberts_2022} also considered that at the infinite width case a fully-trained mean network output is just a linear model based on random features. In this sense, infinite width neural networks are rather shallow in terms of model complexity, however deep they may appear (pg 289). We expect for true emergence of the fundamental laws of physics, one would need an infinitely deep infinite width network.

\item \textbf{Hypergraphs}: It has recently been proposed by \cite{wolfram2020project} that the fundamental laws of physics are explainable by considering our universe as a progression of hypergraph node link updates. Based on the theory of computational equivalence \cite{citeulike:106131}, a neural network and a hypergraph network are all Turing equivalent. Thus, a contracting volume of distributions for a neural network is suggestive that with the progression of time and hypergraph update steps we are collectively honing in towards one final measurement. Let's get it right.

\end{itemize}

\end{document}